\documentclass{article}
\usepackage[margin=1in]{geometry}
\usepackage{graphicx}
\usepackage{natbib}
\usepackage{colortbl}
\usepackage{xcolor}
\usepackage{hhline}
\usepackage{amsmath}
\usepackage{caption}
\usepackage{pdflscape}
\usepackage{rotating}
\usepackage{hyperref}
\usepackage{authblk}
\usepackage{subcaption}
\usepackage{booktabs}
\usepackage{multirow}
\usepackage{longtable}

\title{Benchmarking the Discovery Engine}
\date{June 2025}

\author{Jack Foxabbott\thanks{Corresponding author: jack@leap-labs.com}}
\author{Arush Tagade}
\author{Andrew Cusick}
\author{Robbie McCorkell}
\author{Leo McKee-Reid}
\author{Jugal Patel}
\author{Jamie Rumbelow}
\author{Jessica Rumbelow}
\author{Zohreh Shams}

\affil{Leap Laboratories, London, United Kingdom}

\begin{document}

\maketitle

\begin{abstract}
The Discovery Engine is a general purpose automated system for
scientific discovery, which combines machine learning with state-of-the-art ML interpretability to enable rapid and robust scientific insight across diverse datasets. In this paper, we benchmark the Discovery Engine against five recent peer-reviewed scientific publications applying machine learning across medicine, materials science, social science, and environmental science. In each case, the Discovery Engine matches or exceeds prior predictive performance while also generating deeper, more actionable insights through rich interpretability artefacts. These results demonstrate its potential as a new standard for automated, interpretable scientific modelling that enables complex knowledge discovery from data.
\end{abstract}

\section{Introduction}
Despite major advances in computational power and data availability, the rate of transformative scientific discovery has slowed \citep{bloom2020,chu2021,park2023}. Much of this stagnation stems from limitations in the traditional hypothesis-driven scientific method: it is slow, subject to human cognitive bias, and path-dependent on existing literature. The manual analysis of data remains time-intensive and vulnerable to oversight, particularly in complex, high-dimensional datasets.

Recent developments in automated machine learning (AutoML) have sought to lower the barrier for effective modelling by non-experts. However, many existing systems fall short for scientific applications, in which the objective is to make novel discoveries. They often rely on pre-trained foundation models, which may come with inductive biases that obscure novel findings. Moreover, while some systems offer high predictive performance, they often do so at the expense of interpretability, limiting their usefulness in scientific workflows that demand insight and explanatory power, and preventing their use in safety-critical domains such as health care and aviation.

To address these challenges, we introduce the Discovery Engine \citep{leap2025discovery}, 
a general purpose automated system for scientific discovery. It combines robust model training with novel interpretability methods to provide not just accurate predictions but also meaningful, reproducible explanations of how those predictions arise. The Discovery Engine systematically identifies empirical relationships and outputs structured explanations and pattern artefacts that enable a deeper understanding of patterns present in the dataset.

In this paper, we assess the Discovery Engine’s performance on five diverse datasets drawn from published scientific papers. These datasets span medicine, materials science, environmental science, and social science. For each dataset, we replicate the original modelling pipeline, compare the results to those produced by the Discovery Engine, and analyse the interpretability artefacts generated. We find that our system not only consistently meets or exceeds the original papers’ performance, but also reveals rich, scientifically interesting patterns in the data.

\section{The Discovery Engine}

The Discovery Engine \citep{leap2025discovery} is a general-purpose automated system for
scientific discovery developed at Leap Laboratories for accelerating scientific discovery. It combines machine learning with state-of-the-art ML interpretability to enable rapid and robust scientific insight across diverse datasets.

It was designed in response to growing recognition of the limitations of traditional hypothesis-driven science: that it is slow, biased, and constrained by prior knowledge. The system embodies a new paradigm, data-driven discovery, by autonomously identifying meaningful empirical patterns in complex datasets in the absence of pre-defined hypotheses.

What distinguishes the Discovery Engine from other AutoML platforms is the combination of machine learning and novel interpretability techniques, to both model and explain high-dimensional, non-linear patterns. Unlike traditional statistical methods or pre-trained foundation models (which often require large sample sizes or transfer assumptions from unrelated domains) the Discovery Engine operates efficiently even on small datasets (on the order of hundreds of samples), making it applicable in low-data regimes common across many scientific fields.

\subsection{System Architecture}

The Discovery Engine is composed of four integrated components that together enable end-to-end automated scientific discovery: data preprocessing, automated model training, interpretability and pattern extraction, and structured output generation.

\subsubsection{Data Ingestion and Preprocessing}

Most scientific datasets require extensive cleaning and formatting before they can be used for modelling. The Discovery Engine automates this process entirely by performing intelligent scaling, encoding, imputation, deduplication, and outlier removal based on heuristics derived from the data’s statistical profile. These steps are executed with no manual intervention, allowing domain experts to go from raw data to analysis-ready inputs in minutes. This preprocessing pipeline dramatically reduces the time and expertise required to prepare data for scientific investigation.

\subsubsection{Automated Machine Learning}

The Discovery Engine’s custom AutoML engine explores and automatically selects from a 
wide range of model architectures, including gradient boosting, neural networks, and ensembles. 
The AutoML engine includes robust hyperparameter tuning, overfit detection via hold-out validation, and early stopping. It avoids pre-trained models to ensure all insights are derived solely from the input data, preserving scientific integrity and avoiding confounding from external datasets. This component ensures models are performant and generalisable.

\subsubsection{Interpretability and Pattern Extraction}

Once the best-performing model is selected, the system automatically extracts interpretable patterns. We define a pattern as a set of conditions on a subset of the features that implies something interesting about the target, whether that be increasing or decreasing its mean, increasing the odds of it belonging to a certain class, increasing or decreasing its variance, etc. These patterns are classified as either:

\begin{itemize}
\item \textbf{Discoveries}: patterns with strong empirical support, validated using the available data;
\item \textbf{Hypotheses}: model-inferred patterns that extrapolate from the data and suggest directions for further research.
\end{itemize}

The Discovery Engine’s interpretability stack employs novel techniques to surface robust, nonlinear and combinatorial relationships that are often inaccessible to manual analysis. These insights are grounded in data, not in prior literature, making them especially valuable in domains with sparse or unreliable background knowledge. In this work, we visualise a single pattern for each dataset, but note that Discovery Engine returned many interesting patterns for each. 

\subsubsection{Report Generation and Scientific Outputs}

The Discovery Engine produces a comprehensive suite of outputs designed for use by scientists:

\begin{itemize}
\item A PDF report (also provided as \LaTeX) listing each pattern, with statistical evidence, effect sizes, and novelty rankings;
\item An interactive dashboard for visual exploration of discovered relationships;
\item Exported code and artefacts to enable full reproducibility of the pipeline;
\item LLM-generated summaries that contextualise each validated pattern within existing scientific literature. Language models are used strategically and only after patterns are empirically validated.
\end{itemize}

\section{Benchmark Datasets}

Across the five benchmark datasets, we show that the Discovery Engine not only matches or exceeds the predictive performance of published models, but also yields interpretable artefacts that uncover novel, non-obvious relationships. These artefacts are actionable: they surface new scientific hypotheses, identify boundary conditions, and reveal latent structures with domain-relevant implications.

\begin{enumerate}
    \item \textbf{Hepatitis C Virus (HCV) Prediction} \citep{fan2023ihcp} discussed in Section~\ref{sec:hcv}.
    \item \textbf{Concrete Compressive Strength} \citep{zhang2024predicting} discussed in Section~\ref{sec:ccs}.
    \item \textbf{Climate Change Beliefs} \citep{todorova2025machine} discussed in Section~\ref{sec:ccb}.
    \item \textbf{Ozone Air Quality Prediction} \citep{betancourt2021aq} discussed in Section~\ref{sec:ozone}.
    \item \textbf{Hearing Loss Prediction from NHANES} \citep{MI2025109252} discussed in Section~\ref{sec:hearing-loss}.
\end{enumerate}

Each dataset was automatically preprocessed to match the experimental design of the original paper making performance metrics (e.g., AUC, R\textsuperscript{2}, RMSE) directly comparable. Where applicable, interpretability artefacts were compared qualitatively. A table containing all metrics for the best models reported in the papers and the best models found by Discovery Engine can be found in Appendix~\ref{sec:appendix}. Table~\ref{fig:discovery_comparison} presents the main metric for each target variable that was the basis of evaluation in each respective paper. Since some papers have reported on several targets, we also report on the average performance across all targets for each paper in Table~\ref{fig:discovery_comparison_mean}.
\begin{table}[p]
\footnotesize
\begin{tabular}{lllp{3cm}p{2cm}l c}
\toprule
\textbf{Paper} & \textbf{Section} & \textbf{Target} & \textbf{Trained by} & \textbf{Model} & \textbf{Metric} & \textbf{Value} \\
\midrule

\multirow{2}{*}{\cite{fan2023ihcp}} & \multirow{2}{*}{\hyperref[sec:hcv]{HCV}
} & \multirow{2}{*}{\texttt{Score}} 
& \cite{fan2023ihcp} & Random Forest & Accuracy & 0.915 \\
& & & Discovery Engine & XGBoost & Accuracy & \textbf{0.977} \\
\midrule

\multirow{2}{*}{\cite{zhang2024predicting}} & \multirow{2}{*}{\hyperref[sec:ccs]{CCS}} & \multirow{2}{*}{\texttt{ccs}} 
& \cite{zhang2024predicting} & LightGBM & RMSE & 3.26 \\
& & & Discovery Engine & Neural Net & RMSE & \textbf{0.28} \\
\midrule

\multirow{8}{*}{\cite{todorova2025machine}} & \multirow{8}{*}{\hyperref[sec:ccb]{Climate Beliefs}} & \multirow{2}{*}{\texttt{ccwept}} 
& \cite{todorova2025machine} & GBDT & R\textsuperscript{2} & 0.10 \\
& & & Discovery Engine & Random Forest & R\textsuperscript{2} & \textbf{0.14} \\
\arrayrulecolor{gray}\hhline{~~-----}\arrayrulecolor{black}

& & \multirow{2}{*}{\texttt{ccbelief}} 
& \cite{todorova2025machine} & GBDT & R\textsuperscript{2} & 0.57 \\
& & & Discovery Engine & Random Forest & R\textsuperscript{2} & \textbf{0.63} \\
\arrayrulecolor{gray}\hhline{~~-----}\arrayrulecolor{black}

& & \multirow{2}{*}{\texttt{ccpolicy}} 
& \cite{todorova2025machine} & GBDT & R\textsuperscript{2} & \textbf{0.46} \\
& & & Discovery Engine & Random Forest & R\textsuperscript{2} & 0.44 \\
\arrayrulecolor{gray}\hhline{~~-----}\arrayrulecolor{black}

& & \multirow{2}{*}{\texttt{ccshare}} 
& \cite{todorova2025machine} & GBDT & R\textsuperscript{2} & \textbf{0.74} \\
& & & Discovery Engine & Random Forest & R\textsuperscript{2} & 0.73 \\
\midrule

\multirow{28}{*}{\cite{betancourt2021aq}} & \multirow{28}{*}{\hyperref[sec:ozone]{Ozone}} & \multirow{2}{*}{\texttt{o3\_average\_values}} 
& \cite{betancourt2021aq} & Random Forest & R\textsuperscript{2} & 0.60 \\
& & & Discovery Engine & Neural Net & R\textsuperscript{2} & \textbf{0.67} \\
\arrayrulecolor{gray}\hhline{~~-----}\arrayrulecolor{black}

&  & \multirow{2}{*}{\texttt{o3\_daytime\_avg}} 
& \cite{betancourt2021aq} & Random Forest & R\textsuperscript{2} & 0.63 \\
& & & Discovery Engine & Neural Net & R\textsuperscript{2} & \textbf{0.70} \\
\arrayrulecolor{gray}\hhline{~~-----}\arrayrulecolor{black}

 &  & \multirow{2}{*}{\texttt{o3\_nighttime\_avg}} 
& \cite{betancourt2021aq} & Random Forest & R\textsuperscript{2} & 0.59 \\
& & & Discovery Engine & Neural Net & R\textsuperscript{2} & \textbf{0.71} \\
\arrayrulecolor{gray}\hhline{~~-----}\arrayrulecolor{black}

 &  & \multirow{2}{*}{\texttt{o3\_median}} 
& \cite{betancourt2021aq} & Random Forest & R\textsuperscript{2} & 0.57 \\
& & & Discovery Engine & Neural Net & R\textsuperscript{2} & \textbf{0.74} \\
\arrayrulecolor{gray}\hhline{~~-----}\arrayrulecolor{black}

 &  & \multirow{2}{*}{\texttt{o3\_perc25}} 
& \cite{betancourt2021aq} & Random Forest & R\textsuperscript{2} & 0.63 \\
& & & Discovery Engine & Neural Net & R\textsuperscript{2} & \textbf{0.74} \\
\arrayrulecolor{gray}\hhline{~~-----}\arrayrulecolor{black}

 &  & \multirow{2}{*}{\texttt{o3\_perc75}} 
& \cite{betancourt2021aq} & Random Forest & R\textsuperscript{2} & 0.56 \\
& & & Discovery Engine & Neural Net & R\textsuperscript{2} & \textbf{0.70} \\
\arrayrulecolor{gray}\hhline{~~-----}\arrayrulecolor{black}

 &  & \multirow{2}{*}{\texttt{o3\_perc90}} 
& \cite{betancourt2021aq} & Random Forest & R\textsuperscript{2} & 0.59 \\
& & & Discovery Engine & Neural Net & R\textsuperscript{2} & \textbf{0.64} \\
\arrayrulecolor{gray}\hhline{~~-----}\arrayrulecolor{black}

 &  & \multirow{2}{*}{\texttt{o3\_perc98}} 
& \cite{betancourt2021aq} & Random Forest & R\textsuperscript{2} & 0.59 \\
& & & Discovery Engine & Neural Net & R\textsuperscript{2} & \textbf{0.68} \\
\arrayrulecolor{gray}\hhline{~~-----}\arrayrulecolor{black}

 &  & \multirow{2}{*}{\texttt{o3\_dma8eu}} 
& \cite{betancourt2021aq} & Random Forest & R\textsuperscript{2} & 0.58 \\
& & & Discovery Engine & Neural Net & R\textsuperscript{2} & \textbf{0.68} \\
\arrayrulecolor{gray}\hhline{~~-----}\arrayrulecolor{black}

 &  & \multirow{2}{*}{\texttt{o3\_avgdma8epax}} 
& \cite{betancourt2021aq} & Random Forest & R\textsuperscript{2} & 0.63 \\
& & & Discovery Engine & Neural Net & R\textsuperscript{2} & \textbf{0.69} \\
\arrayrulecolor{gray}\hhline{~~-----}\arrayrulecolor{black}

 &  & \multirow{2}{*}{\texttt{o3\_drmdmax1h}} 
& \cite{betancourt2021aq} & Random Forest & R\textsuperscript{2} & 0.51 \\
& & & Discovery Engine & Neural Net & R\textsuperscript{2} & \textbf{0.62} \\
\arrayrulecolor{gray}\hhline{~~-----}\arrayrulecolor{black}

 & & \multirow{2}{*}{\texttt{o3\_w90}} 
& \cite{betancourt2021aq} & Random Forest & R\textsuperscript{2} & 0.51 \\
& & & Discovery Engine & Neural Net & R\textsuperscript{2} & \textbf{0.75} \\
\arrayrulecolor{gray}\hhline{~~-----}\arrayrulecolor{black}

& & \multirow{2}{*}{\texttt{o3\_aot40}} 
& \cite{betancourt2021aq} & Random Forest & R\textsuperscript{2} & 0.60 \\
& & & Discovery Engine & Neural Net & R\textsuperscript{2} & \textbf{0.62} \\
\arrayrulecolor{gray}\hhline{~~-----}\arrayrulecolor{black}

 & & \multirow{2}{*}{\texttt{o3\_nvgt070}} 
& \cite{betancourt2021aq} & Neural Net & R\textsuperscript{2} & \textbf{0.32} \\
& & & Discovery Engine & Neural Net & R\textsuperscript{2} & 0.24 \\
\arrayrulecolor{gray}\hhline{~~-----}\arrayrulecolor{black}

 & & \multirow{2}{*}{\texttt{o3\_nvgt100}} 
& \cite{betancourt2021aq} & Neural Net & R\textsuperscript{2} & 0.12 \\
& & & Discovery Engine & Neural Net & R\textsuperscript{2} & \textbf{0.61} \\
\midrule

\multirow{2}{*}{\cite{MI2025109252}} & \multirow{2}{*}{\hyperref[sec:hearing-loss]{Hearing Loss}} & \multirow{2}{*}{\texttt{HL}} 
& \cite{MI2025109252} & Random Forest & Accuracy & 0.891 \\
& & & Discovery Engine & Random Forest & Accuracy & \textbf{0.893} \\

\bottomrule
\end{tabular}
\caption{Comparison of Discovery Engine with best model in each paper for each target, showing the key metric with the best value in bold.}
\label{fig:discovery_comparison}
\end{table}

\begin{table}[p]
\footnotesize
\centering
\begin{tabular}{lccc}
\toprule
\textbf{Paper} & \textbf{Their mean score} & \textbf{Our mean score} & \textbf{Metric} \\
\midrule
\cite{fan2023ihcp}          & 0.915 & \textbf{0.977} & Accuracy (higher is better) \\
\cite{zhang2024predicting} & 3.260 & \textbf{0.280} & RMSE (lower is better) \\
\cite{todorova2025machine} & 0.468 & \textbf{0.485} & R\textsuperscript{2} (higher is better) \\
\cite{betancourt2021aq}  & 0.535 & \textbf{0.653} & R\textsuperscript{2} (higher is better) \\
\cite{MI2025109252}          & 0.891   & \textbf{0.893}   & Accuracy (higher is better) \\
\bottomrule
\end{tabular}
\caption{Performance of the best models from Discovery Engine and each paper, averaged over all targets.}
\label{fig:discovery_comparison_mean}
\end{table}

\subsection{Hepatitis C Virus Prediction}\label{sec:hcv}

\cite{fan2023ihcp} introduces interpretable hepatitis C prediction (IHCP), an interpretable system for hepatitis C virus (HCV) diagnostic prediction based on black-box machine learning models. Using serological blood test data, the authors train multiple classifiers and select a Bayesian-optimised random forest model \citep{snoek2012practical} as their best performer. This model achieves impressive generalisation on independent test data, with accuracy, precision, recall, and F1-score all exceeding 90\%. To interpret this model, the study combines global feature attributions via SHAP \citep{lundberg2017unified} with a stability-enhanced variant of LIME \citep{ribeiro2016should} for local explanations. The interpretability results reveals that, globally AST (Aspartate Aminotransferase), GGT (Gamma-Glutamyl Transferase), and ALP (Alkaline Phosphatase) emerge as the three most important biomarkers for HCV prediction.

\begin{figure}[h]
    \centering
    \includegraphics[width=0.9\linewidth]{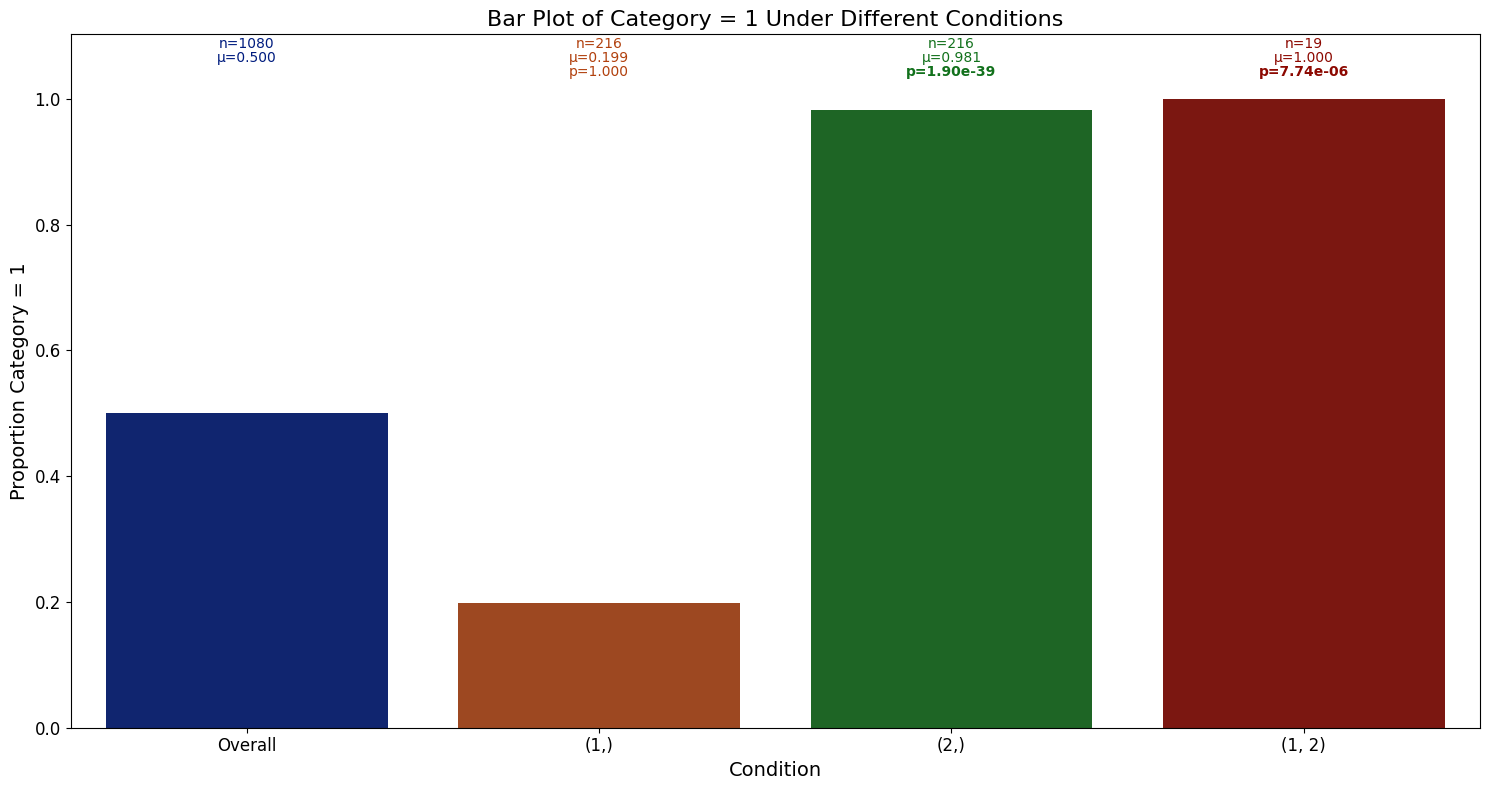}
    \caption{
        Proportion of positive HCV diagnoses under the following conditions. Above each bar we display the following: $n$, the number of samples satisfying the conditions; $\mu$, the average target value under these conditions; $p$, the p-value from a one-tailed proportions z-test comparing the target distribution under the given conditions with the overall target distribution. \\
        (1) ALP between 50 and 66.\\
        (2) AST in the top 20\%.
    }
    \label{fig:hcv_pattern}
\end{figure}

 As shown in Appendix~\ref{sec:appendix}, our XGBoost classifier exceeds the paper’s random forest model across all reported metrics except AUC, for which it comes close. Discovery Engine's interpretability analysis further confirms the clinical relevance of AST, GGT, and ALP as the most globally important features, but also reveals more nuanced patterns. For instance, Figure~\ref{fig:hcv_pattern} illustrates a compound condition that strongly predicts HCV positivity: AST values in the top 20\% of the sample, when combined with ALP values in the range 50–66 IU/L. Interestingly, ALP in this range alone is associated with lower risk, but in the presence of high AST, the risk sharply increases. HCV infection is known to consistently elevate AST levels \citep{Huynh2018HCV, Schreiner2020HCV}, but ALP is only associated with HCV infection for values greater than 120 IU/L \citep{Bagheri2024HCV}, pointing to a potential unknown interaction between these two indicators. This is an example of a non-additive, non-linear interaction that would be difficult to detect using traditional statistical models or single-feature attributions, and underscores Discovery Engine’s potential for surfacing complex diagnostic rules in medical datasets.



\subsection{Concrete Compressive Strength}\label{sec:ccs}

The Nature paper ~\cite{zhang2024predicting} presents an interpretable framework for predicting the compressive strength of high-performance concrete (HPC) using machine learning. The authors evaluate four model families (Random Forest \citep{breiman2001random}, AdaBoost \citep{freund1995desicion}, XGBoost \citep{chen2016xgboost}, and LightGBM \citep{ke2017lightgbm}) with hyperparameters tuned via random search. Their best-performing model, an ensemble-based LightGBM referred to as FR\_LightGBM, achieves superior accuracy compared to previous approaches, reaching an RMSE of 3.26~MPa on the test set. Beyond predictive performance, they employ SHAP \citep{lundberg2017unified} to interpret both global and local feature importance, identifying age, water/cement ratio, slag content, and water as the most influential features.

Using the same dataset, the Discovery Engine’s neural network model significantly improves upon the RMSE and MAE reported in the original study. As illustrated in Figure~\ref{fig:discovery_comparison}, the neural network achieves the lowest error across both metrics, outperforming FR\_LightGBM results reported in the paper by an order of magnitude. This suggests the neural network may be capturing more complex relationships in the data, particularly non-linear and interaction effects that may be missed by a simpler model.

\begin{figure}[h]
    \centering
    \includegraphics[width=0.85\linewidth]{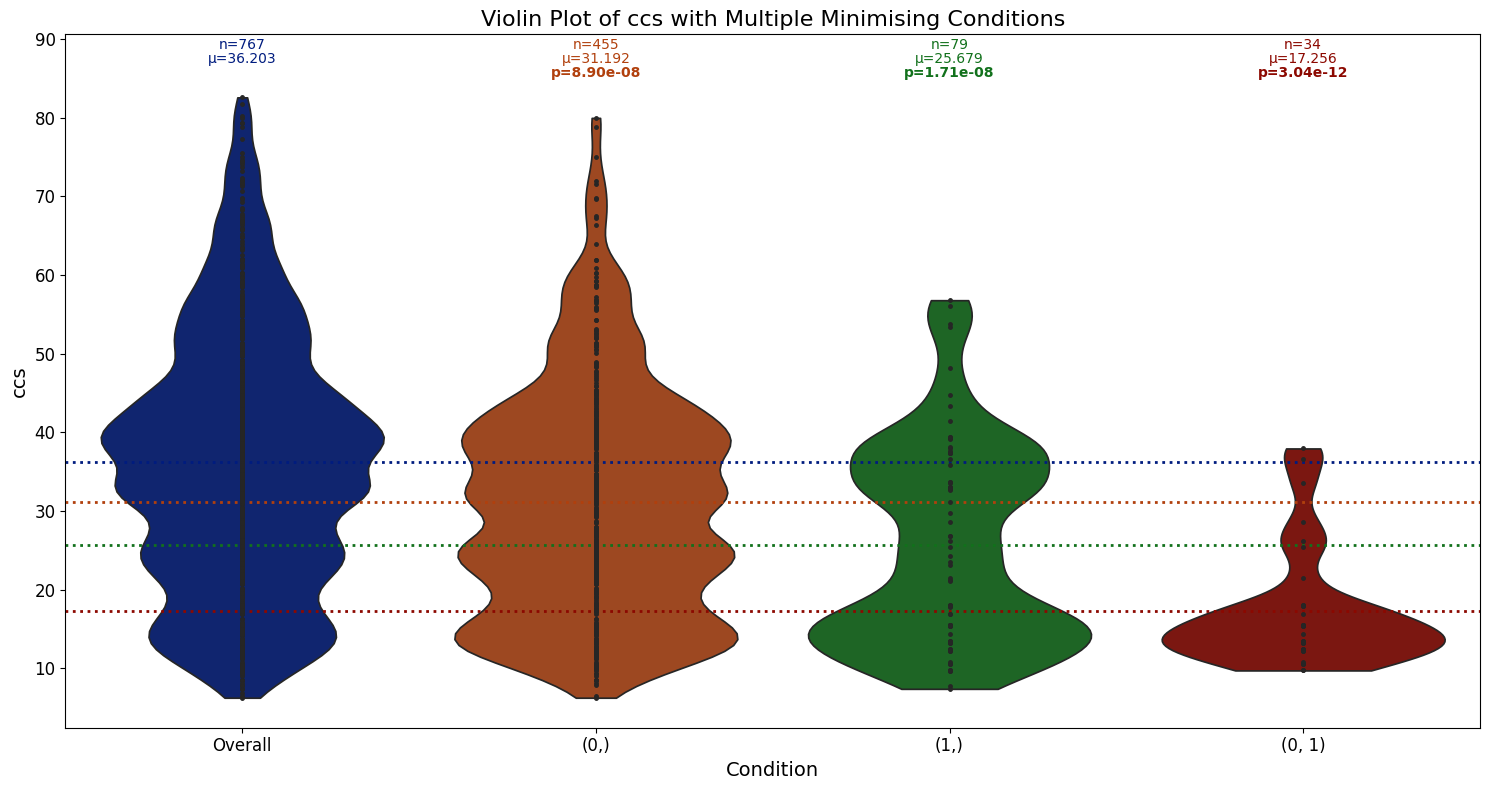}
    \caption{Violin plot of compressive strength (CCS) under the following conditions. Above each bar we display the following: $n$, the number of samples satisfying the conditions; $\mu$, the average target value under these conditions; $p$, the p-value from a one-tailed Mann-Whitney test comparing the target distribution under the given conditions with the overall target distribution.
    \\ (0) Blast Furnace Slag in the bottom 10\%. 
    \\ (1) Aggregate Cement Ratio between 9.9 and 11.1.
    }
    \label{fig:minimising_ccs}
\end{figure}

To interpret these gains, we used the Discovery Engine’s interpretability module to compute global feature importances and extract high-impact patterns. Unlike the SHAP analysis in the original paper, the Discovery Engine highlights the aggregate/cement ratio, coarse-to-fine aggregate ratio, and blast furnace slag as the three most influential variables. Since our neural network achieves higher performance, it has learnt to represent patterns not captured by the authors' FR\_LightGBM, so it is unsurprising that the most important features differ between the two. One such pattern involving the features most important to our neural network, shown in Figure~\ref{fig:minimising_ccs}, reveals that compressive strength is significantly reduced when (1) blast furnace slag content is in the bottom decile and (2) the aggregate/cement ratio lies between 9.9 and 11.1. This pattern aligns with established material science principles \citep{blastfurnaceslag, kozul1997effects}.


\subsection{Climate Change Beliefs}\label{sec:ccb}

A recent paper ~\citep{todorova2025machine} applies interpretable machine learning to investigate which individual- and nation-level factors best predict climate-related beliefs and behaviours across a globally diverse sample. Using gradient-boosted decision trees (GBDTs) trained on responses from 4,635 participants across 55 countries, the authors model four key outcomes: belief in anthropogenic climate change (\texttt{ccbelief}), support for climate policy (\texttt{ccpolicy}), willingness to share pro-climate information on social media (\texttt{ccshare}), and participation in an environmentally beneficial but effortful behavioural task (\texttt{ccwept}). Their GBDT models achieve strong performance for belief ($R^2 = 0.57$), policy support ($R^2 = 0.46$), and social media sharing (74\% classification accuracy), but substantially lower for the behavioural task ($R^2 = 0.10$).

The authors then use SHAP to estimate feature importance across all four outcomes. Four features— environmentalist identity, trust in climate science, internal environmental motivation, and national Human Development Index (HDI)—consistently emerge as the most predictive. However, the strength and direction of influence vary significantly across outcomes and national contexts, highlighting the complex interplay between individual psychology and structural context in shaping climate-relevant behaviour.

\begin{figure}[h]
    \centering
    \includegraphics[width=0.85\linewidth]{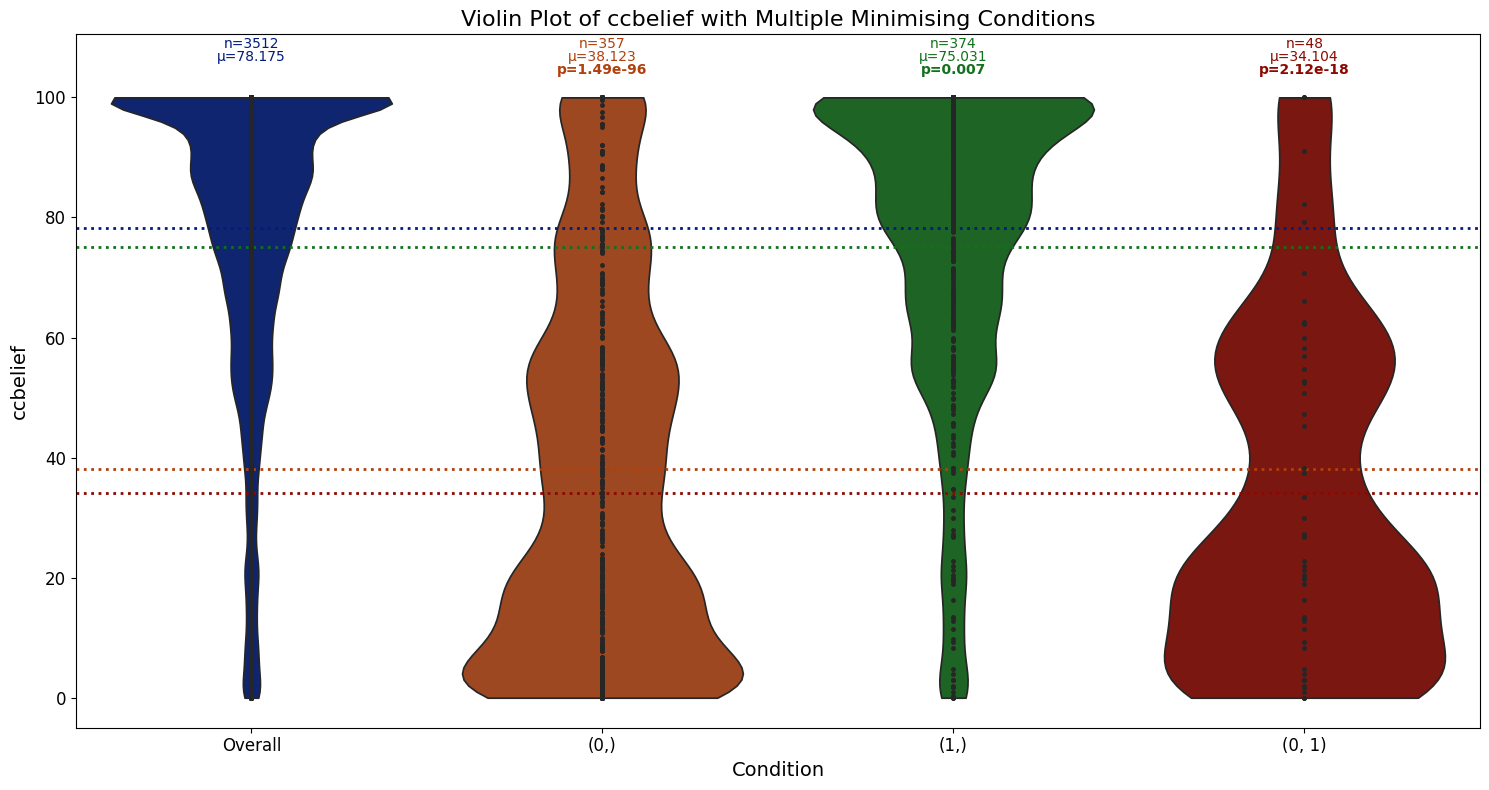}
    \caption{Violin plot of climate belief under the following conditions. Above each bar we display the following: $n$, the number of samples satisfying the conditions; $\mu$, the average target value under these conditions; $p$, the p-value from a one-tailed Mann-Whitney test comparing the target distribution under the given conditions with the overall target distribution. 
    \\(0) Trust in Science in the bottom decile. 
    \\(1) Climate Risk Index in the top decile.
    }
    \label{fig:cc_pattern}
\end{figure}

The Discovery Engine’s random forest models surpass the performance of the paper’s GBDT baselines on average, as shown in Figure~\ref{fig:discovery_comparison}, and identify 3 of the 4 features quoted in the paper as most important (external environmental motivation being ranked higher than HDI). In addition, Discovery Engine’s pattern extraction revealed interpretable multi-factor interactions. One notable pattern, illustrated in Figure~\ref{fig:cc_pattern}, shows that belief in climate change is particularly low among participants with low trust in science \textit{and} who reside in countries among the top 10\% of the Climate Risk Index \citep{Germanwatch2025CRI}. This hypothesis is evidenced by some academic works, showing that individuals living in higher risk areas of Canada \citep{pickering2016barriers} and Australia \citep{Tranter2023Trust} are more climate skeptical, but this pattern points to a broader trend that deserves further investigation.


\subsection{Ozone Air Quality Prediction}\label{sec:ozone}
\begin{figure}[h]
    \centering
    \includegraphics[width=0.85\linewidth]{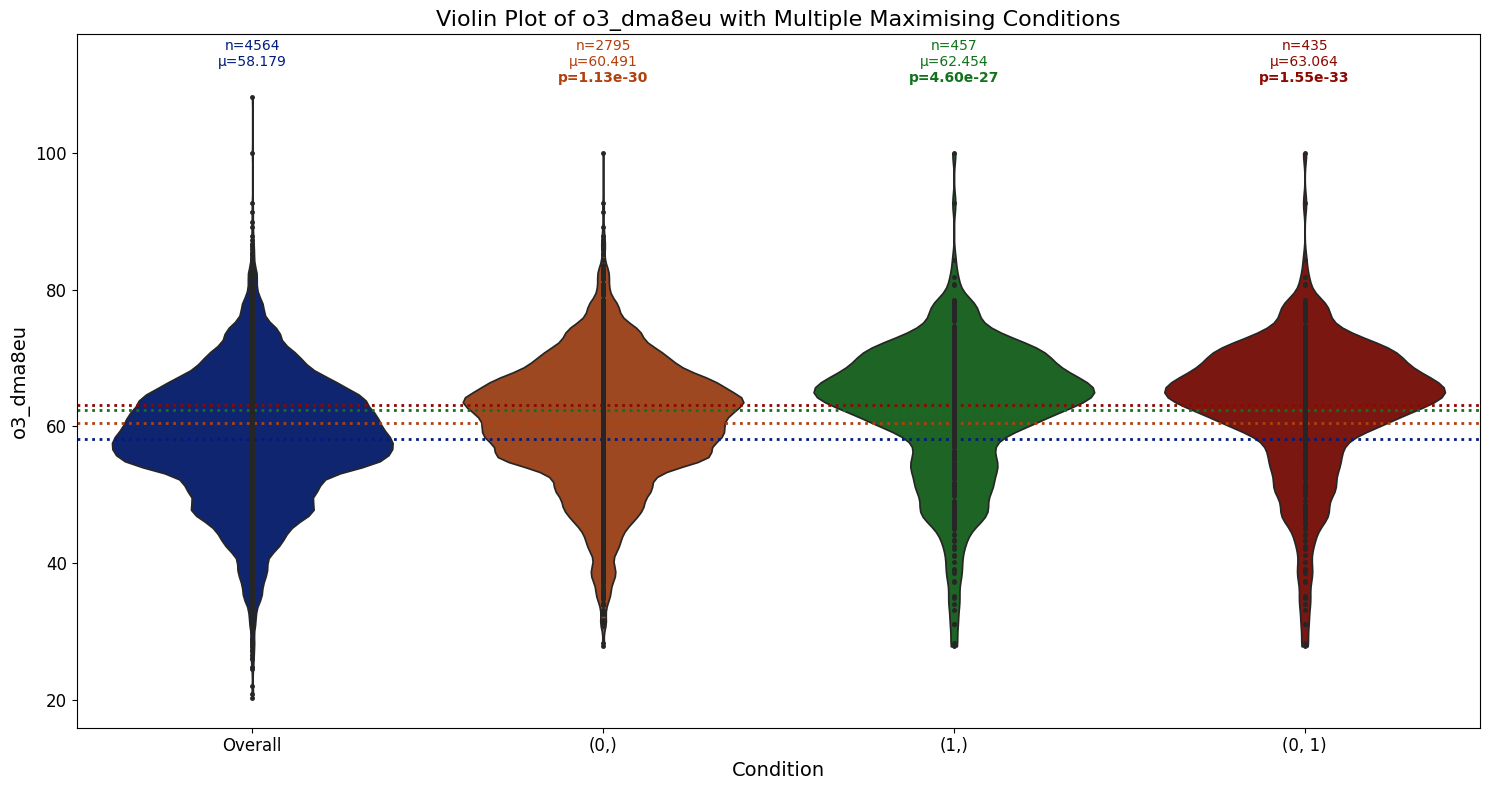}
    \caption{Violin plot of ozone levels (dma8eu) under the following conditions. Above each bar we display the following: $n$, the number of samples satisfying the conditions; $\mu$, the average target value under these conditions; $p$, the p-value from a one-tailed Mann-Whitney test comparing the target distribution under the given conditions with the overall target distribution.
    \\ (0) Climatic zone is warm.
    \\ (1) Nitrous oxide column density in the top decile.}
    
    \label{fig:ozone_pattern}
\end{figure}

The AQ-Bench dataset ~\citep{betancourt2021aq} provides a globally harmonised benchmark for machine learning on tropospheric ozone. It aggregates five years of ozone metrics (2010–2014) across over 5500 monitoring stations, paired with rich station-level metadata reflecting environmental, socio-economic, and climatic influences. AQ-Bench is structured to support supervised learning tasks that map these metadata to long-term ozone metrics, and includes baseline evaluations using linear regression, random forest, and shallow neural networks. These baselines establish a performance benchmark for predictive modelling and help identify where more sophisticated techniques might yield improvements, particularly for ozone extremes, which are both scientifically important and technically challenging to predict.

In the original study, random forests achieved the highest $R^2$ scores across most ozone metrics, with performance typically exceeding 50\% for core targets such as daytime average ozone and AOT40. These results underscore the complexity of tropospheric ozone formation, which involves nonlinear, multi-scale interactions between emissions, climate, and geography.

Leap’s Discovery Engine outperformed the best baseline model on 14 out of 15 ozone targets. As shown in Figure~\ref{fig:discovery_comparison}, its neural networks consistently achieved higher $R^2$ scores than the original baselines, particularly on high-impact metrics such as the daily maximum 8-hour average ozone (\texttt{dma8eu}) and vegetation exposure indices (\texttt{AOT40}, \texttt{W90}). Importantly, the Discovery Engine also surfaced interpretable, empirically grounded patterns predictive of ozone extremes. One such pattern, shown in Figure~\ref{fig:ozone_pattern}, reveals that when the climatic zone has Köppen climate classification \citep{Koppen1936} \texttt{warm moist} or \texttt{warm dry}, and the nitrous oxide column density falls in the top 10\% of observed values, the \texttt{dma8eu} metric is significantly elevated. Existing literature explains this pattern by enhanced photochemical efficiency under warm conditions, with ozone production rates increase more rapidly than ozone destruction rates \citep{meng2023chemical, acp-16-11601-2016}. This kind of multi-factor relationship illustrates the Discovery Engine’s ability not just to improve prediction, but to support mechanistic understanding and data-driven hypothesis generation in atmospheric science.


\subsection{Hearing Loss Prediction (NHANES)}\label{sec:hearing-loss}
\begin{figure}[p]
    \begin{subfigure}[b]{\textwidth}
        \centering
        \includegraphics[width=\linewidth]{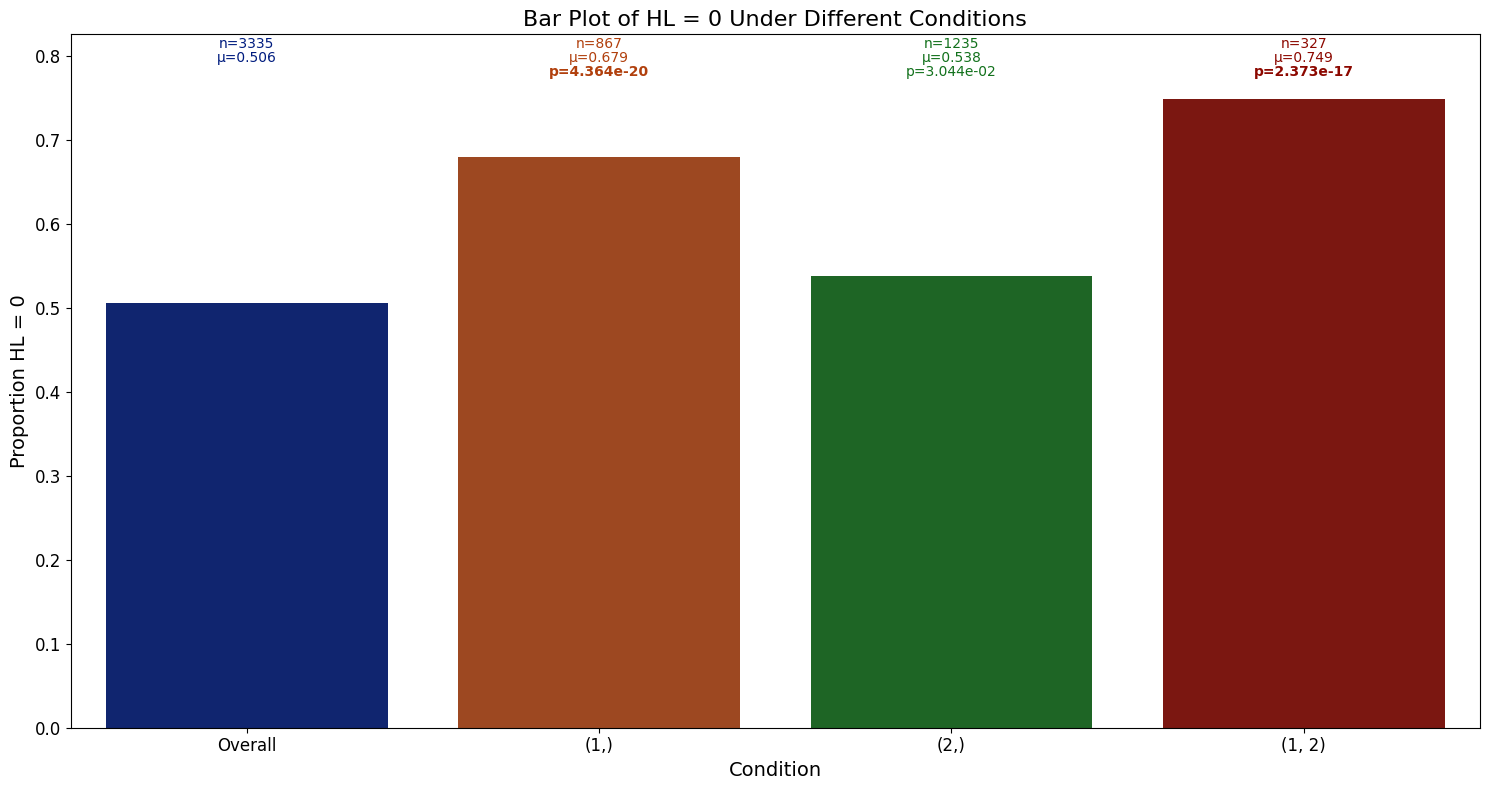}
        \caption{Bar plot of fraction of Hearing Loss (HL) = 0 under the following conditions:
        \\(1) Age pointing to middle-aged patients
        \\(2) Vitamin E intake in the top decile}
        \label{fig:hl-pat-vite}
    \end{subfigure}
    \begin{subfigure}[b]{\textwidth}
        \centering
        \includegraphics[width=\linewidth]{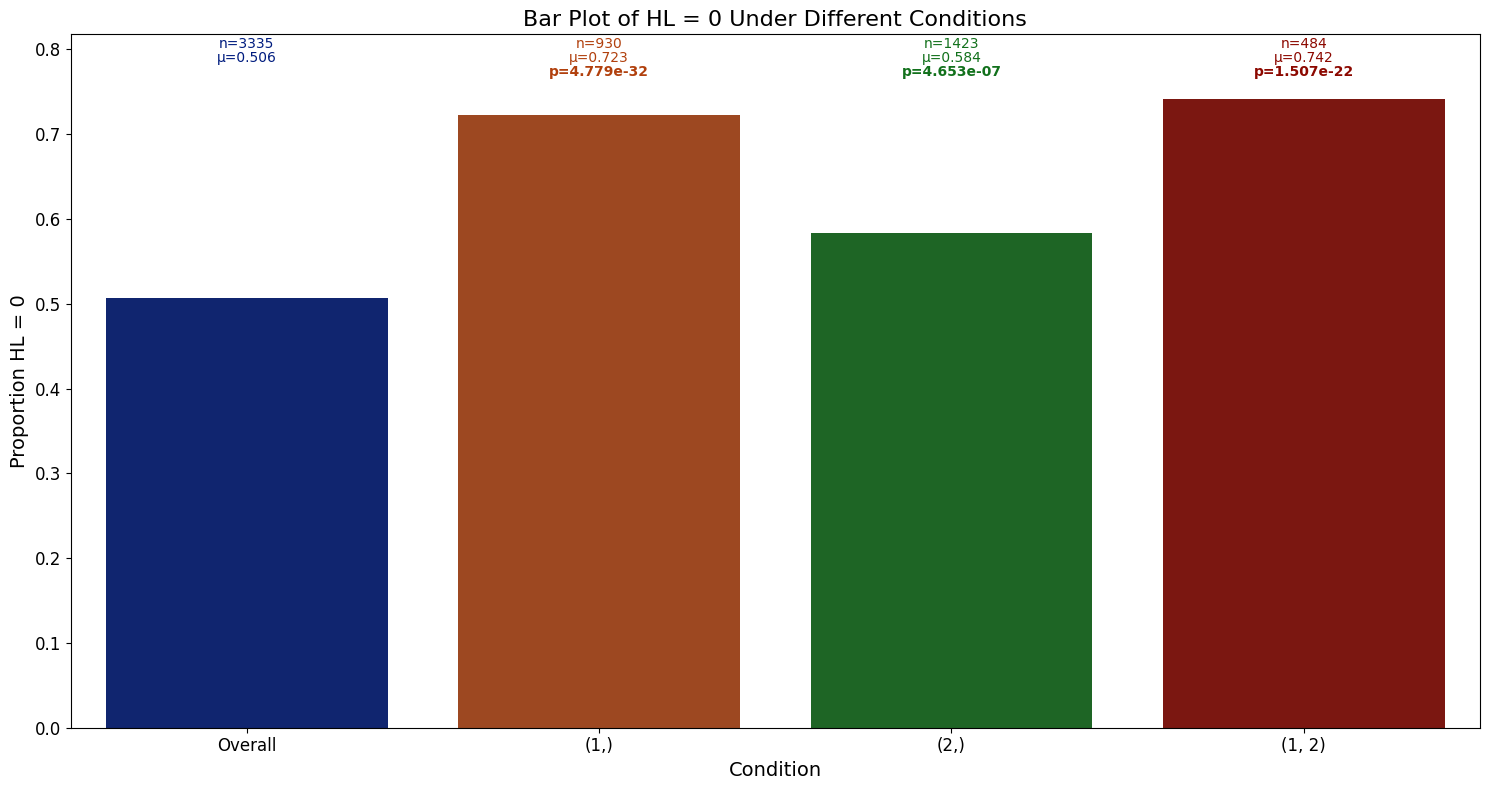}
        \caption{Bar plot of fraction of Hearing Loss (HL) = 0 under the following conditions:
        \\ (1) Age pointing to middle-aged patients
        \\ (2) Lead content in blood being in the bottom decile}
        \label{fig:hl-pat-bloodpb}
    \end{subfigure}
    \caption{Patterns extracted by the Discovery Engine showing inputs that minimise risk of hearing loss. Above each bar we display the following: $n$, the number of samples satisfying the conditions; $\mu$, the average target value under these conditions; $p$, the p-value from a one-tailed proportions z-test comparing the target distribution under the given conditions with the overall target distribution.}
    \label{fig:hl-pat}
\end{figure}

The NHANES dataset~\citep{eke2012prevalence}, contains demographic, clinical, and laboratory data from 2003 to 2018, and has been used to investigate factors associated with hearing loss. Building on this resource, a recent study by \cite{MI2025109252} selected a subset of 2,814 participants aged 18–60 who had complete hearing test results and comprehensive blood and urine metal measurements. The authors trained several supervised learning models, including Random Forests, XGBoost, Support Vector Machines, and fully connected neural networks, to predict the probability of hearing loss.

They found that their Random Forest model performed the best in terms of accuracy and their feature importance analysis identified \texttt{age}, \texttt{blood\_Pb} (i.e lead concentration in blood), and \texttt{vitamin\_E} levels as the most influential predictors. Their multivariate analysis revealed statistically significant associations between elevated \texttt{blood\_Pb} and low \texttt{vitamin\_E} intake with an increased risk of hearing loss.

Our Discovery Engine Random Forest model achieves competitive performance as shown in Figure \ref{fig:discovery_comparison} and reproduced similar patterns, as illustrated in Figure \ref{fig:hl-pat}. Specifically, we observed that reduced \texttt{blood\_Pb} levels and higher \texttt{vitamin\_E} intake, especially among middle-aged individuals, were associated with a lower predicted risk of hearing loss. These replicated associations are supported by past work~\citep{liu2018hearing, gonccalves2024effectiveness} and highlight the potential of automated pattern discovery using the Discovery Engine in health-related datasets.



\newpage
\section{Conclusion}
Our results confirm that the Discovery Engine offers a robust, general-purpose solution to automated scientific modelling. It consistently matches or surpasses benchmarked models from five peer-reviewed scientific studies, while providing more interpretable results. These interpretability artefacts are not only useful for explaining model behaviour; they also serve as sources of new empirical insight, revealing patterns that are non-obvious and often non-linear. Moreover, it does this completely automatically. The papers replicated here represent weeks or months of work by the original authors, while we completed the analysis in a few hours. The Discovery Engine has the potential to massively accelerate the research cycle, while reducing reliance on hypothesis-driven workflows, and mitigating many of the biases that have historically hindered scientific progress.

By reducing the cognitive and technical burden of data analysis, the Discovery Engine makes exploratory modelling and data-driven discovery accessible to a wider range of scientists. We believe that this tool represents a critical step forward in the pursuit of open, interpretable, and efficient science, and sets a precedent for what automated systems can achieve in advancing human knowledge.

\newpage
\bibliographystyle{plainnat}
\bibliography{refs.bib}

\newpage
\appendix
\section{Appendix}\label{sec:appendix}
\begin{sidewaystable}
\scriptsize
\begin{longtable}{lllp{3cm}p{2cm}cccccccc}
\toprule
\textbf{Paper} & \textbf{Section} & \textbf{Target} & \textbf{Trained by} & \textbf{Model} & \multicolumn{8}{c}{\textbf{Metric}} \\
\cmidrule(lr){6-13}
& & & & & \textbf{Accuracy} & \textbf{F1 Score} & \textbf{Precision} & \textbf{Recall} & \textbf{AUC} & \textbf{RMSE} & \textbf{R\textsuperscript{2}} & \textbf{MAE} \\
\midrule
\endfirsthead

\toprule
\textbf{Paper} & \textbf{Section} & \textbf{Target} & \textbf{Trained by} & \textbf{Model} & \multicolumn{8}{c}{\textbf{Metric}} \\
\cmidrule(lr){6-13}
& & & & & \textbf{Accuracy} & \textbf{F1 Score} & \textbf{Precision} & \textbf{Recall} & \textbf{AUC} & \textbf{RMSE} & \textbf{R\textsuperscript{2}} & \textbf{MAE} \\
\midrule
\endhead

\multirow{2}{*}{\cite{fan2023ihcp}} & \multirow{2}{*}{\hyperref[sec:hcv]{HCV}} & \texttt{Score} & \cite{fan2023ihcp} & Random Forest & 0.915 & 0.905 & 0.901 & 0.923 & \textbf{0.990} & --- & --- & --- \\
& & \texttt{Score} & Discovery Engine & XGBoost & \textbf{0.977} & \textbf{0.977} & \textbf{0.967} & \textbf{0.983} & 0.977 & --- & --- & --- \\
\midrule
\multirow{2}{*}{\cite{zhang2024predicting}} & \multirow{2}{*}{\hyperref[sec:ccs]{CCS}} & \texttt{ccs} & \cite{zhang2024predicting} & LightGBM & --- & --- & --- & --- & --- & 3.26 & --- & 2.35 \\
& & \texttt{ccs} & Discovery Engine & Neural Net & --- & --- & --- & --- & --- & \textbf{0.28} & --- & \textbf{0.21} \\
\midrule
\multirow{8}{*}{\cite{todorova2025machine}} & \multirow{8}{*}{\hyperref[sec:ccb]{Climate Beliefs}} & \texttt{ccwept} & \cite{todorova2025machine} & GBDT & --- & --- & --- & --- & --- & --- & 0.1 & --- \\
& & \texttt{ccwept} & Discovery Engine & Random Forest & --- & --- & --- & --- & --- & --- & \textbf{0.14} & --- \\
\arrayrulecolor{gray}
\hhline{~~-----------}
\arrayrulecolor{black}
& & \texttt{ccbelief} & \cite{todorova2025machine} & GBDT & --- & --- & --- & --- & --- & --- & 0.57 & --- \\
& & \texttt{ccbelief} & Discovery Engine & Random Forest & --- & --- & --- & --- & --- & --- & \textbf{0.63} & --- \\
\arrayrulecolor{gray}
\hhline{~~-----------}
\arrayrulecolor{black}
& & \texttt{ccpolicy} & \cite{todorova2025machine} & GBDT & --- & --- & --- & --- & --- & --- & \textbf{0.46} & --- \\
& & \texttt{ccpolicy} & Discovery Engine & Random Forest & --- & --- & --- & --- & --- & --- & 0.44 & --- \\
\arrayrulecolor{gray}
\hhline{~~-----------}
\arrayrulecolor{black}
& & \texttt{ccshare} & \cite{todorova2025machine} & GBDT & --- & --- & --- & --- & --- & --- & \textbf{0.74} & --- \\
& & \texttt{ccshare} & Discovery Engine & Random Forest & --- & --- & --- & --- & --- & --- & 0.73 & --- \\

\midrule
\multirow{28}{*}{\cite{betancourt2021aq}} & \multirow{28}{*}{\hyperref[sec:ozone]{Ozone}} & \texttt{o3\_average\_values} & \cite{betancourt2021aq} & Random Forest & --- & --- & --- & --- & --- & --- & 0.60 & --- \\
& & \texttt{o3\_average\_values} & Discovery Engine & Neural Net & --- & --- & --- & --- & --- & --- & \textbf{0.67} & --- \\
\arrayrulecolor{gray}
\hhline{~~-----------}
\arrayrulecolor{black}
& & \texttt{o3\_daytime\_avg} & \cite{betancourt2021aq} & Random Forest & --- & --- & --- & --- & --- & --- & 0.63 & --- \\
& & \texttt{o3\_daytime\_avg} & Discovery Engine & Neural Net & --- & --- & --- & --- & --- & --- & \textbf{0.70} & --- \\
\arrayrulecolor{gray}
\hhline{~~-----------}
\arrayrulecolor{black}
& & \texttt{o3\_nighttime\_avg} & \cite{betancourt2021aq} & Random Forest & --- & --- & --- & --- & --- & --- & 0.59 & --- \\
& & \texttt{o3\_nighttime\_avg} & Discovery Engine & Neural Net & --- & --- & --- & --- & --- & --- & \textbf{0.71} & --- \\
\arrayrulecolor{gray}
\hhline{~~-----------}
\arrayrulecolor{black}
& & \texttt{o3\_median} & \cite{betancourt2021aq} & Random Forest & --- & --- & --- & --- & --- & --- & 0.57 & --- \\
& & \texttt{o3\_median} & Discovery Engine & Neural Net & --- & --- & --- & --- & --- & --- & \textbf{0.74} & --- \\
\arrayrulecolor{gray}
\hhline{~~-----------}
\arrayrulecolor{black}
& & \texttt{o3\_perc25} & \cite{betancourt2021aq} & Random Forest & --- & --- & --- & --- & --- & --- & 0.63 & --- \\
& & \texttt{o3\_perc25} & Discovery Engine & Neural Net & --- & --- & --- & --- & --- & --- & \textbf{0.74} & --- \\
\arrayrulecolor{gray}
\hhline{~~-----------}
\arrayrulecolor{black}
& & \texttt{o3\_perc75} & \cite{betancourt2021aq} & Random Forest & --- & --- & --- & --- & --- & --- & 0.56 & --- \\
& & \texttt{o3\_perc75} & Discovery Engine & Neural Net & --- & --- & --- & --- & --- & --- & \textbf{0.70} & --- \\
\arrayrulecolor{gray}
\hhline{~~-----------}
\arrayrulecolor{black}
& & \texttt{o3\_perc90} & \cite{betancourt2021aq} & Random Forest & --- & --- & --- & --- & --- & --- & 0.59 & --- \\
& & \texttt{o3\_perc90} & Discovery Engine & Neural Net & --- & --- & --- & --- & --- & --- & \textbf{0.64} & --- \\
\arrayrulecolor{gray}
\hhline{~~-----------}
\arrayrulecolor{black}
& & \texttt{o3\_perc98} & \cite{betancourt2021aq} & Random Forest & --- & --- & --- & --- & --- & --- & 0.59 & --- \\
& & \texttt{o3\_perc98} & Discovery Engine & Neural Net & --- & --- & --- & --- & --- & --- & \textbf{0.68} & --- \\
\arrayrulecolor{gray}
\hhline{~~-----------}
\arrayrulecolor{black}
& & \texttt{o3\_dma8eu} & \cite{betancourt2021aq} & Random Forest & --- & --- & --- & --- & --- & --- & 0.58 & --- \\
& & \texttt{o3\_dma8eu} & Discovery Engine & Neural Net & --- & --- & --- & --- & --- & --- & \textbf{0.68} & --- \\
\arrayrulecolor{gray}
\hhline{~~-----------}
\arrayrulecolor{black}
& & \texttt{o3\_avgdma8epax} & \cite{betancourt2021aq} & Random Forest & --- & --- & --- & --- & --- & --- & 0.63 & --- \\
& & \texttt{o3\_avgdma8epax} & Discovery Engine & Neural Net & --- & --- & --- & --- & --- & --- & \textbf{0.69} & --- \\
\arrayrulecolor{gray}
\hhline{~~-----------}
\arrayrulecolor{black}
& & \texttt{o3\_drmdmax1h} & \cite{betancourt2021aq} & Random Forest & --- & --- & --- & --- & --- & --- & 0.51 & --- \\
& & \texttt{o3\_drmdmax1h} & Discovery Engine & Neural Net & --- & --- & --- & --- & --- & --- & \textbf{0.62} & --- \\
\arrayrulecolor{gray}
\hhline{~~-----------}
\arrayrulecolor{black}
& & \texttt{o3\_w90} & \cite{betancourt2021aq} & Random Forest & --- & --- & --- & --- & --- & --- & 0.51 & --- \\
& & \texttt{o3\_w90} & Discovery Engine & Neural Net & --- & --- & --- & --- & --- & --- & \textbf{0.75} & --- \\
\arrayrulecolor{gray}
\hhline{~~-----------}
\arrayrulecolor{black}
& & \texttt{o3\_aot40} & \cite{betancourt2021aq} & Random Forest & --- & --- & --- & --- & --- & --- & 0.60 & --- \\
& & \texttt{o3\_aot40} & Discovery Engine & Neural Net & --- & --- & --- & --- & --- & --- & \textbf{0.62} & --- \\
\arrayrulecolor{gray}
\hhline{~~-----------}
\arrayrulecolor{black}
& & \texttt{o3\_nvgt070} & \cite{betancourt2021aq} & Neural Net & --- & --- & --- & --- & --- & --- & \textbf{0.32} & --- \\
& & \texttt{o3\_nvgt070} & Discovery Engine & Neural Net & --- & --- & --- & --- & --- & --- & 0.24 & --- \\
\arrayrulecolor{gray}
\hhline{~~-----------}
\arrayrulecolor{black}
& & \texttt{o3\_nvgt100} & \cite{betancourt2021aq} & Neural Net & --- & --- & --- & --- & --- & --- & 0.12 & --- \\
& & \texttt{o3\_nvgt100} & Discovery Engine & Neural Net & --- & --- & --- & --- & --- & --- & \textbf{0.61} & --- \\
\midrule
\multirow{2}{*}{\cite{MI2025109252}} & \multirow{2}{*}{\hyperref[sec:hearing-loss]{Hearing Loss}} &  \texttt{HL} & \cite{MI2025109252} & Random Forest & 0.891 & 0.881 & \textbf{0.896} & 0.912 & \textbf{0.947} & --- & --- & --- \\
& & \texttt{HL} & Discovery Engine & Random Forest & \textbf{0.893} & \textbf{0.892} & 0.855 & \textbf{0.925} & 0.895 & --- & --- & --- \\
\bottomrule
\end{longtable}
\end{sidewaystable}

\end{document}